\documentclass[accepted]{uai2026}

\usepackage[american]{babel}
\usepackage{amsmath}
\usepackage{graphicx}
\usepackage{booktabs}
\usepackage{dblfloatfix}
\usepackage{caption}
\usepackage{algorithm}
\usepackage{algpseudocode}
\usepackage{tcolorbox}
\usepackage{comment}
\usepackage{amssymb,url}

\usepackage{natbib} 
\usepackage{times}

\title{No More Maybe-Arrows: \\Resolving Causal Uncertainty by Breaking Symmetries}

\author{\href{mailto:<t.huang2@student.tue.nl>?Subject=Your paper}{Tingrui Huang}{}}
\author{Devendra Singh Dhami}
\affil{%
    Department of Mathematics and Computer Science\\
    Eindhoven University of Technology
}

\begin{document}

\maketitle
\thispagestyle{numbered} 
\pagestyle{numbered}

\begin{abstract}
The recent works on causal discovery have followed a similar trend of learning partial ancestral graphs (PAGs) since observational data constrain the true causal directed acyclic graph (DAG) only up to a Markov equivalence class. This limits their application in the majority of downstream tasks, as uncertainty in causal relations remains unresolved. We propose a new refinement framework, CausalSAGE, for converting PAGs to DAGs while respecting the underlying causal relations. The framework expands discrete variables into state-level representations, constrains the search space using structural knowledge and soft priors, and applies a unified differentiable objective for joint optimization. The final DAG is obtained by aggregating the optimized structures and enforcing acyclicity when necessary. Our experimental evaluations show that the obtained DAGs preserve the underlying causal relations while also being efficient to obtain.
\end{abstract}

\section{Introduction}
The fact that humans reason causally is very well studied in the literature from several perspectives, such as philosophical~\citep{hall1934time,white1990ideas,williamson2007causality}, social science~\citep{marini1988causality}, and cognitive science~\citep{weichwald2021causality,gerstenberg2024counterfactual}. While it is relatively simpler for humans, understanding cause–effect relationships poses a far more difficult challenge for machines~\citep{pylyshyn1980causal,bishop2021artificial}. One of the major reasons behind this is the lack of interventional and counterfactual data, as such experimental interventions are costly, unethical, or even infeasible. Thus, in several high impact domains, ranging from biology to economics, researchers seek to move beyond predictive associations and uncover the underlying causal structure from observational data that is more readily available.  Consequently, a lot of research in this field focuses on learning the causal graph from observational data, a task referred to as causal discovery~\citep{spirtes2016causal,zanga2022survey}.

\begin{figure}[t]
    \centering
    \includegraphics[width=\linewidth]{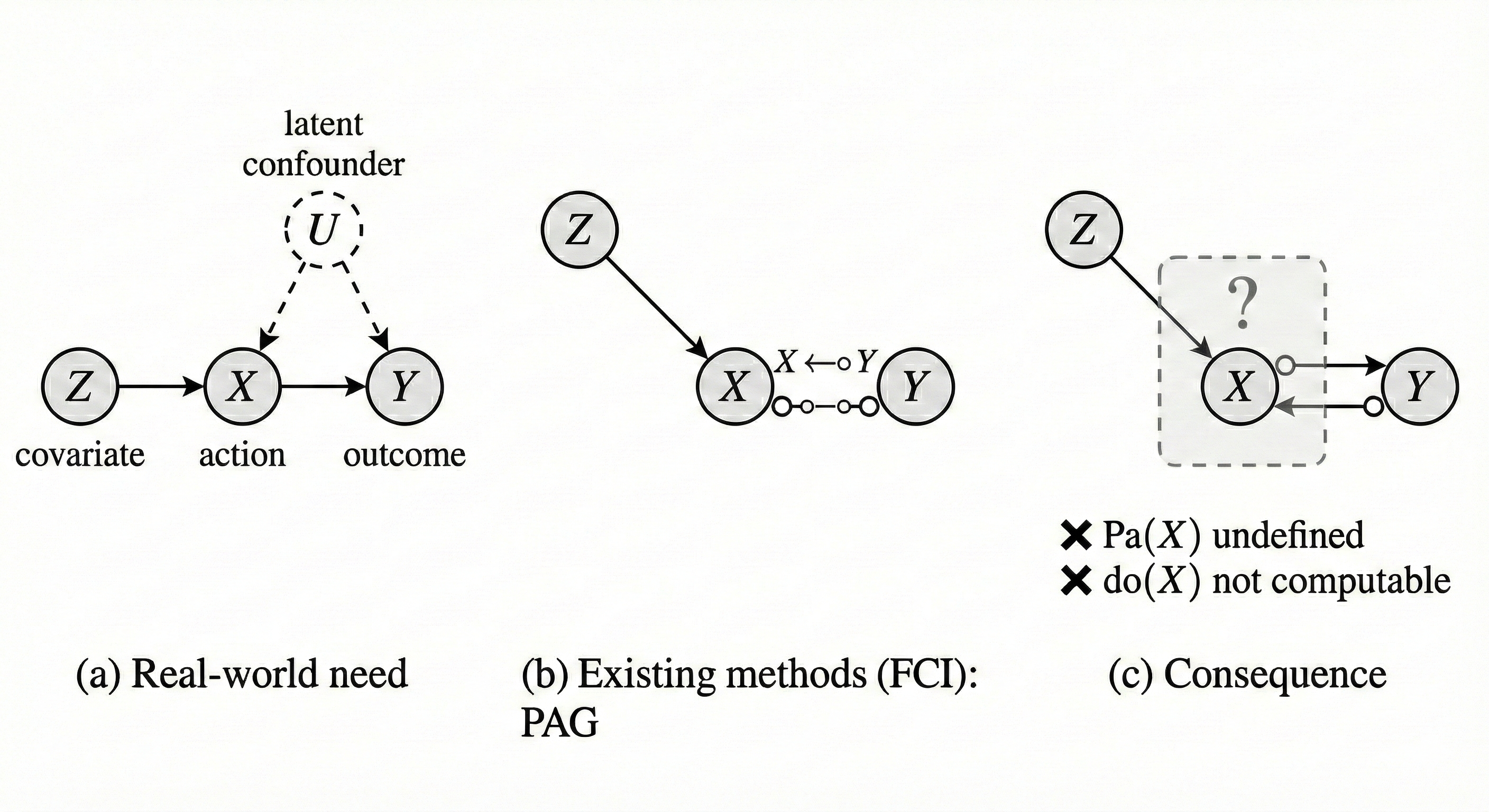}
    \caption{\textbf{Why PAGs are not enough.}
(a) Most downstream tasks require a fully specified causal DAG.
(b) FCI handles latent confounders but outputs partially oriented graphs (PAGs).
(c) Ambiguous parent sets (e.g., $\mathrm{Pa}(X)$) prevent computing interventional effects.}
    \label{fig:motivation}
\end{figure}

Causal discovery algorithms can be classified into two major categories: score-based and constraint-based. Score-based algorithms search for the most probable causal structure by assessing goodness-of-fit scores, such as the Bayesian Information Criterion (BIC), for different possible structures. Examples include greedy equivalence search (GES)~\citep{chickering2002optimal} and LiNGAM~\citep{shimizu2006linear}. Constraint-based algorithms construct a causal structure that aligns with all observed conditional independencies,identified using conditional independence tests. Examples include the Peter-Clark (PC) algorithm~\citep{spirtes2000causation,kalisch2007estimating} and fast causal inference (FCI)~\citep{spirtes1993discovery,spirtes2001anytime,colombo2012learning}. Under standard assumptions such as the Markov assumption and faithfulness condition, observational data constrain the true causal directed acyclic graph (DAG) only up to a Markov equivalence class. Constraint-based and hybrid algorithms, such as those based on Fast Causal Inference, therefore return a partial ancestral graph (PAG) rather than a fully oriented DAG. A PAG compactly represents an equivalence class of causal graphs that encodes the same conditional independence relations over the observed variables while explicitly capturing uncertainty about edge orientations and latent confounding. 

While PAGs represent causal uncertainty in a robust manner, their disadvantage is that they leave unresolved ambiguity, as strict causal directions cannot be inferred. See figure~\ref{fig:motivation} for a visual representation. The challenge, therefore, is not only to learn a PAG from data but also to systematically refine it by resolving orientation ambiguities in a principled manner.

We propose a conversion of PAGs to DAGs through a 3-stage differentiable refinement framework we call Causal \textit{S}tate-\textit{A}ugmented \textit{G}raph \textit{E}stimation (CausalSAGE~\footnote{We make our code publicly available at \url{https://github.com/tingrui-huang/causal-sage}}). First, we perform a state-level expansion in which each discrete variable is represented using a one-hot encoding, enabling fine-grained structural modeling at the level of individual states. We then restrict the search space by imposing hard structural constraints derived from the PAG skeleton and its identified v-structures, thereby ensuring consistency with the equivalence class while incorporating soft asymmetry-inducing priors. Finally, we formulate a unified differentiable objective that jointly optimizes data reconstruction likelihood and structural regularization, allowing principled and efficient end-to-end learning of a DAG consistent with the input PAG. Our experimental evaluations on benchmarks ranging from small (11 nodes) to large (724 nodes) show that we are able to efficiently convert PAGs to DAGs while respecting the underlying causal relations.

We proceed as follows: we first touch upon the related works before explaining our refinement procedure in detail. We then present our extensive empirical evaluations, demonstrating both the efficacy and effectiveness of our method before concluding.


\section{Related Work}
Breaking the Markov equivalence class while trying to learn causal relations from observational data has been studied in the literature~\citep{shimizu2006linear,brouillard2020differentiable,van2023beyond}, but it generally requires extra assumptions~\citep{perkovic2017interpreting}. Thus, causal discovery has become one of the most studied topics~\citep{spirtes2016causal,nogueira2022methods,zheng2024causal}. 

With the advent of large language models in recent years, a major focus has shifted to refining the obtained causal graph in order to be closer to the ground truth graph wherever available~\citep{ankan2025expert,febrinanto2025refined,zhang2025causalchat}. Although this line of research remains in an early stage, its practical relevance is evident. Refining a PAG with multiple undirected or ambiguously oriented edges into a fully specified DAG yields a model with explicit causal semantics, thereby facilitating estimation and downstream inference.


\section{CausalSAGE}

We propose a differentiable refinement framework that transforms a partially oriented PAG into a fully directed DAG.
The framework consists of three tightly coupled stages: (1) State-level expansion transforms discrete variables into one-hot state representations.
(2) Hard structural constraints derived from the PAG skeleton and v-structures define the feasible adjacency space, while soft priors introduce controlled asymmetry to avoid symmetric saddle points.
(3) A unified differentiable objective jointly optimizes reconstruction likelihood and structural regularization.
The final DAG is obtained by aggregating optimized adjacency matrices and enforcing acyclicity if necessary.

\subsection{State-Aware Representation}

Most discrete causal discovery methods treat each variable as a single categorical node. All its internal states are aggregated, and causal directions are decided only at this coarse variable level.
However, different states of a variable may interact with other variables in very different ways. For example, state $a_1$ of $V_i$ may consistently lead to state $b_2$ of $V_j$, while other states show no such pattern.

Recent work shows that explicitly modeling the internal state structure of discrete variables can reveal additional asymmetry in the data-generating process \citep{ni2022ordinalcausaldiscovery}. 
Motivated by this insight, we adopt a state-aware representation that models interactions at the level of individual variable states rather than aggregated variable units.

\vspace{0.5em}

\paragraph{State Expansion.}
Let $V_i$ be a discrete variable with $L_i$ possible states. 
We expand $V_i$ into a one-hot vector
\[
\mathbf{z}_i \in \{0,1\}^{L_i}, \quad \|\mathbf{z}_i\|_1 = 1,
\]
Let $\mathcal I_i = \{1,\dots,L_i\}$ denote the index set of states for variable $V_i$ where each component corresponds to one specific state of $V_i$. 
For a system with $M$ variables, the full state representation is obtained by concatenation:
\[
\mathbf{z} = [\mathbf{z}_1^\top, \mathbf{z}_2^\top, \dots, \mathbf{z}_M^\top]^\top 
\in \mathbb{R}^{\sum_{i=1}^{M} L_i}.
\]
This transformation preserves state-level information that is collapsed under variable-level modeling.

\vspace{0.5em}

\paragraph{Block-wise Parameterization.}
Based on the expanded state representation, we introduce a global weight matrix
\[
\mathbf{W} \in \mathbb{R}^{\left(\sum_{i=1}^{M} L_i\right) \times \left(\sum_{i=1}^{M} L_i\right)}.
\]
The matrix $\mathbf{W}$ is naturally partitioned into sub-matrices (blocks)
\[
\mathbf{W}_{ij} \in \mathbb{R}^{L_i \times L_j},
\]
where each block $\mathbf{W}_{ij}$ parameterizes the influence from variable $V_i$ to variable $V_j$ at the state level. 
Each element within $\mathbf{W}_{ij}$ captures a dependency from one specific state of $V_i$ to one specific state of $V_j$.

We treat each block $\mathbf{W}_{ij}$ as the fundamental unit of variable-level causality. 
Directional decisions between $V_i$ and $V_j$ are obtained by aggregating the learned state-level parameters within $\mathbf{W}_{ij}$.

\subsection{Structural Encoding of PAG}
\label{sec:pag_encoding}

The input PAG defines structural feasibility constraints encoded directly in the parameter space before optimization.

\paragraph{Variable-level admissibility.}
For each variable pair $(V_i, V_j)$, we define a directional admissibility indicator $m_{ij}\in\{0,1\}$ according to PAG semantics:
(i) if $V_i$ and $V_j$ are non-adjacent, both directions are forbidden;
(ii) if the direction $V_i \to V_j$ is resolved, only $i\to j$ is admissible;
(iii) otherwise, both directions remain admissible.
Additionally, unshielded colliders implied by the PAG are preserved by forbidding reverse directions of forced parents.

\paragraph{State-level lifting.}
Variable-level admissibility is lifted to the state level by expanding each admissible direction to its full state block. We denote the resulting state-level mask as $\mathbf S\in\{0,1\}^{n_s\times n_s}$,
where
\[
S_{ab} = m_{ij}, \qquad a\in\mathcal I_i,\; b\in\mathcal I_j.
\]
The final learnable adjacency is then
\[
\mathbf A = \sigma(\mathbf W)\odot \mathbf S,
\]
ensuring that all structurally forbidden connections remain identically zero throughout training.

\begin{figure*}
    \centering
    \includegraphics[width=0.8\textwidth]{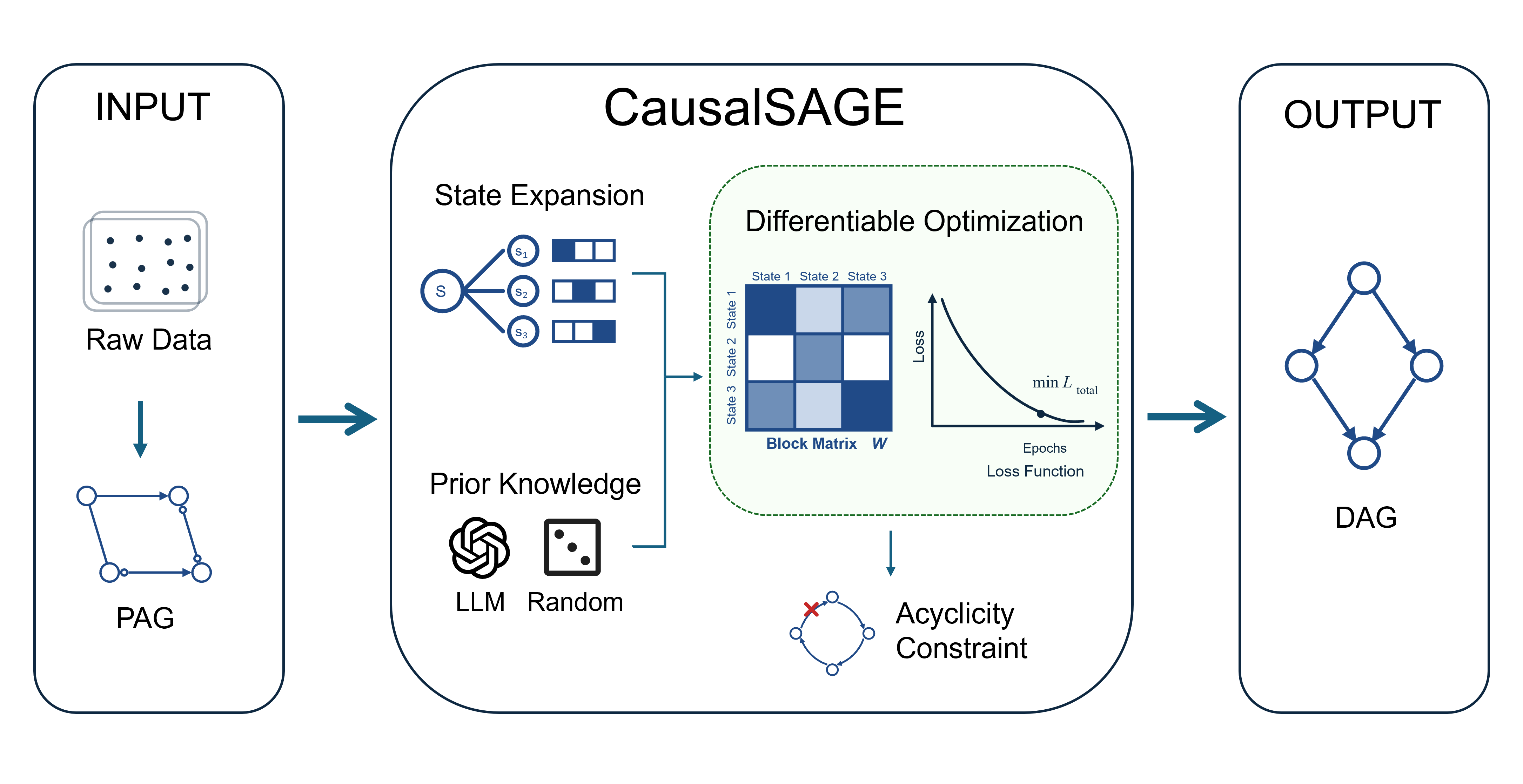}
    \caption{\textbf{CausalSAGE}. The framework expands discrete variables into state representations, constrains the search space using random/LLM priors, and finally applies a unified differentiable objective for joint optimization to output a DAG.}
    \label{fig:arch}
\end{figure*}

\subsection{Likelihood-Based Direction Selection}
\label{sec:dir_softmax}

\paragraph{Reconstruction signal.}
For edges whose direction is \emph{unresolved} in the input PAG,
we rely on a data-driven signal derived from a shared state-level reconstruction objective.
All candidate orientations are optimized jointly under the same likelihood term.
Directional preference emerges implicitly:
if the block $\mathbf{W}_{ij}$ contributes more to reducing the reconstruction loss,
its parameters receive larger gradient updates,
while the reverse block is suppressed by structural regularization.
Fully resolved edges are hard-constrained by the mask and do not participate in directional competition.

\paragraph{State-level logit construction.}
Let $\mathbf{X}\in\{0,1\}^{N\times n_s}$ be the expanded state data matrix (one-hot states concatenated between variables), and let $\mathbf{S}\in\{0,1\}^{n_s\times n_s}$ be the hard mask induced by the PAG skeleton (state-level allowed connections).
We maintain a trainable logit matrix $\mathbf{W}\in\mathbb{R}^{n_s\times n_s}$, and construct masked logits
\begin{equation}
\label{eq:masked_logits}
\mathbf{L} \;=\; \mathbf{X}\bigl(\mathbf{W}\odot \mathbf{S}\bigr),
\end{equation}
where $\odot$ denotes element-wise multiplication.
Intuitively, $\mathbf{S}$ prevents the model from using state-to-state connections that are disallowed by the input PAG.

\paragraph{Block strength aggregation (for edge interpretation and extraction).}
Let $\mathcal{U}$ denote the set of \emph{unresolved} variable pairs in the input PAG.
For $(i,j)\in\mathcal{U}$,
let $\mathcal I_i$ and $\mathcal I_j$ denote the state index sets of variables $V_i$ and $V_j$.
After optimization, we summarize the learned directional strength of a candidate orientation by aggregating the corresponding state-level block:
\begin{equation}
\label{eq:block_strength_max}
s_{i\to j}
\;=\;
\max_{a\in \mathcal I_i,\; b\in \mathcal I_j}
\Bigl(A_{ab}\Bigr),
\qquad
\text{where } \mathbf{A}=\sigma(\mathbf{W})\odot \mathbf{S}.
\end{equation}
This aggregation is used in our evaluation pipeline to extract a variable-level edge $V_i\to V_j$ by thresholding $s_{i\to j}$.
Directional asymmetry is not enforced by an explicit two-model comparison 
but arises from the shared reconstruction objective, together with
cycle and sparsity regularization, that discourage symmetric solutions.

\paragraph{Cross-entropy reconstruction objective.}
Given the masked logits in \eqref{eq:masked_logits}, we obtain predicted state probabilities $\hat{\mathbf{X}}$ by applying a per-variable softmax over the state set of each variable.\footnote{Concretely, for each variable $V_v$, the softmax is restricted to its state index set $\mathcal I_v$.}
We define the predicted probabilities by a block-wise softmax:
\begin{equation}
\hat{X}_{n,s}
=
\frac{\exp(L_{n,s})}
{\sum_{s' \in \mathcal I_v} \exp(L_{n,s'})},
\quad s \in \mathcal I_v.
\end{equation}
The reconstruction loss is written as the standard cross-entropy between observed one-hot states and predicted probabilities:
\begin{equation}
\label{eq:recon_ce}
\mathcal{L}_{\text{recon}}
\;=\;
-\frac{1}{NM}\sum_{n=1}^{N}\sum_{v=1}^{M}\;\sum_{s\in \mathcal I_v}
X_{n,s}\,\log \hat{X}_{n,s},
\end{equation}
where $\mathcal I_v$ denotes the state index set of variable $V_v$ and the softmax normalization is applied within each $S_v$.

In our implementation, $\mathcal{L}_{\text{recon}}$ is computed over all variables; restricting the reconstruction to variables incident to unresolved pairs is a straightforward variant.

This objective encourages each variable to be reconstructed from its allowed incoming state connections,
so candidate orientations compete through their contribution to this shared likelihood term.

\subsection{Structural Regularization}
While $\mathcal{L}_{\text{recon}}$ determines which direction better explains the data,
it alone does not eliminate weak connections or suppress bidirectional activations.
We therefore introduce three structural regularizers with complementary roles:
block-wise sparsity removes weak blocks,
the cycle penalty discourages simultaneous activation of both directions,
and the skeleton-preservation term keeps adjacencies identified by the input PAG from being overly reduced during optimization.

\paragraph{Block-wise Sparsity via Weighted Group Lasso ($\mathcal{L}_{\text{sparse}}$):}
State-level expansion substantially enlarges the parameter space,
and unresolved pairs initially allow both directional blocks.
To suppress weak or redundant block activations and avoid dense bidirectional structures, we impose sparsity at the block level.

Let $\mathcal B$ denote the set of directed variable pairs
for which the corresponding state-level block is allowed by the mask.
Each block $\mathbf A_{ij}$ represents the state-to-state interactions
from $V_i$ to $V_j$. 

We apply a weighted group-lasso penalty:
\begin{equation}
\label{eq:sparse_loss}
\mathcal{L}_{\text{sparse}}
=
\frac{1}{|\mathcal{B}|}
\sum_{(i,j)\in\mathcal{B}}
\left\|
\mathbf{A}_{ij}\odot\mathbf{P}_{ij}
\right\|_F.
\end{equation}
where $\mathbf P_{ij}\in\mathbb{R}_+^{L_i\times L_j}$ is a non-negative penalty matrix,
and $\|\cdot\|_F$ denotes the Frobenius norm.

This formulation shrinks entire state-to-state blocks jointly, encouraging edge-level sparsity. 
The penalty matrix $\mathbf P_{ij}$ compensates for state-imbalance effects.
Without weighting, dominant baseline states may disproportionately influence
block norms, potentially obscuring informative but less frequent interactions.

\paragraph{Pairwise Directional Suppression ($\mathcal{L}_{\text{cycle}}$):}
To prevent simultaneous activation of both directions within a variable pair, we introduce a block-level directional suppression term.

Let $\mathcal Q$ denote the set of unordered variable pairs
for which both directions are structurally admissible.

For each unordered variable pair $\{i,j\}\in\mathcal{Q}$,
we define the directional block norms:
\[
c_{i\to j}=\|\mathbf{A}_{ij}\|_F,
\qquad
c_{j\to i}=\|\mathbf{A}_{ji}\|_F.
\]

The cycle penalty is defined as
\begin{equation}
\mathcal{L}_{\text{cycle}}
=
\frac{1}{|\mathcal{Q}|}
\sum_{\{i,j\}\in\mathcal{Q}}
c_{i\to j}\,c_{j\to i}.
\end{equation}

If both directional blocks have large norms, the product term becomes large,
penalizing bidirectional activation.
If one direction is suppressed, the penalty vanishes.
This regularizer therefore promotes asymmetric edge selection
without imposing hard acyclicity constraints. A detailed cycle analysis is provided in Appendix ~\ref{sec:cycle_analysis}

\paragraph{Skeleton Preservation ($\mathcal{L}_{\text{skeleton}}$):}

While sparsity and directional suppression encourage edge pruning, their combined effect may lead to unintended removal of both directions for edges that are supported by the input PAG skeleton. To prevent such collapse, we introduce a skeleton-preservation term. Specifically, for entries allowed by the skeleton mask $\mathbf{S}$, we penalize the simultaneous disappearance of both directions by encouraging the bidirectional sum to remain non-zero:

\begin{equation}
\label{eq:skeleton_loss}
\mathcal{L}_{\text{skeleton}}
=
\frac{1}{\|\mathbf{S}\|_1}
\sum_{a,b}
\mathbf{S}[a,b]\,
\Bigl(1-\bigl(A_{ab}+A_{ba}\bigr)\Bigr)^2.
\end{equation}

This term does not enforce bidirectionality.
Instead, it prevents skeleton-supported edges from being entirely pruned,
while allowing $\mathcal{L}_{\text{recon}}$ and $\mathcal{L}_{\text{cycle}}$
to determine the final orientation.

\begin{algorithm}[t]
\caption{PAG-to-DAG Refinement via Differentiable Optimization}
\label{alg:refinement}
\begin{algorithmic}[1]
\Require Discrete data $\mathcal{D}$ over $\mathcal{V}$, input PAG $\mathcal{P}$, $\lambda_1,\lambda_2,\lambda_3$, threshold $\tau$, steps $T$, step size $\eta$
\Ensure Refined DAG $\mathcal{G}=(\mathcal{V},\mathcal{E})$
\State One-hot expand $\mathcal{D}\!\to\!\mathbf{X}\in\{0,1\}^{N\times n_s}$ with state sets $\{\mathcal{I}_i\}$, and encode $\mathcal{P}\!\to\!\mathbf{S}\in\{0,1\}^{n_s\times n_s}$
\State Initialize logits $\mathbf{W}_0$ with soft prior bias on unresolved pairs and mask $\mathrm{diag}(\mathbf{W}_0)=-\infty$
\For{$t=0$ to $T-1$}
\State $\mathbf{W}_{t+1}\gets \mathbf{W}_t-\eta\nabla_{\mathbf{W}_t}\Big[\mathcal{L}_{\mathrm{recon}}(\mathbf{X},\sigma(\mathbf{W}_t)\odot\mathbf{S})+\lambda_1\mathcal{L}_{\mathrm{sparse}}+\lambda_2\mathcal{L}_{\mathrm{cycle}}+\lambda_3\mathcal{L}_{\mathrm{skel}}\Big]$
\EndFor
\State $\mathbf{A}\gets\sigma(\mathbf{W}_T)\odot\mathbf{S}$ and $s_{ij}\gets\max_{a\in\mathcal{I}_i,b\in\mathcal{I}_j}A_{ab}$
\State $\mathcal{E}\gets\{i\!\to\!j\mid s_{ij}>\tau,\, i\neq j\}$ and break cycles by iteratively removing the minimum-$s_{ij}$ edge on a detected cycle
\State \Return $\mathcal{G}=(\mathcal{V},\mathcal{E})$
\end{algorithmic}
\end{algorithm}

\subsection{Unified Differentiable Objective}

After introducing all components, we combine them into a unified optimization framework.

The overall objective is defined as:
\begin{equation}
\label{eq:total_loss}
\mathcal{L}_{\text{total}}
=
\mathcal{L}_{\text{recon}}
+
\lambda_1 \mathcal{L}_{\text{sparse}}
+
\lambda_2 \mathcal{L}_{\text{cycle}}
+
\lambda_3 \mathcal{L}_{\text{skeleton}}.
\end{equation}

Here, $\mathcal{L}_{\text{recon}}$ corresponds to the conditional likelihood objective full state-expanded data. Directional preference emerges implicitly from joint optimization. Implementation and optimization details are provided in Appendix~\ref{sec:setup_details}.

$\mathcal{L}_{\text{sparse}}$ enforces block-level sparsity, 
$\mathcal{L}_{\text{cycle}}$ discourages simultaneous activation of opposite directions, 
and $\mathcal{L}_{\text{skeleton}}$ preserves adjacencies identified in the initial PAG skeleton.

The hyperparameters $\lambda_1, \lambda_2, \lambda_3$ balance data fidelity and structural regularization.

After training, we obtain $\mathbf W^\star$ and the corresponding continuous adjacency
\[
\mathbf A^\star=\sigma(\mathbf W^\star)\odot\mathbf S.
\]
A variable-level graph is extracted by block-max thresholding,
as detailed in Sec.~\ref{sec:final_dag_check}.

\subsection{Prior-Based Symmetry Breaking}
Although the reconstruction term provides directional gradients and the cycle penalty prevents bidirectional edges, their interaction can create a mutual balancing effect for unresolved pairs within an MEC. 
Both orientations may initially receive comparable gradient signals, while the cycle penalty suppresses simultaneous growth. 
As a result, the system can settle into a balanced state where neither direction clearly dominates. 
A detailed gradient analysis is provided in Appendix~\ref{app:symmetry_analysis}.
To resolve this directional ambiguity, we introduce a soft symmetry-breaking prior over unresolved variable pairs. 
Unlike the hard structural mask, this prior does not forbid any admissible direction; instead, it provides a asymmetric bias at initialization, allowing optimization to escape the mutual balancing described above.

\paragraph{Random prior.}
The primary role of the prior is to introduce asymmetry in unresolved variable pairs. A simple instantiation is to assign a randomly chosen directional bias to each unresolved pair, resulting in asymmetric initialization of the corresponding logit blocks. 

\paragraph{LLM-based semantic prior.}
As an alternative instantiation, when variable names carry meaningful semantics, an LLM-based prior can provide a context-aware directional initialization. 
For each unresolved pair $(V_i,V_j)$, we query GPT-3.5-turbo to estimate directional plausibility and translate the resulting preference into a biased initialization of the corresponding logit blocks. 
This formulation retains the same symmetry-breaking mechanism while optionally incorporating semantic cues.
Implementation details are provided in Appendix~\ref{prior_details}.

\paragraph{Strong initialization.}
The prior is implemented as an asymmetric initialization of the corresponding logit blocks. 
Specifically, we assign initial probabilities of 0.9/0.1 to the favored and unfavored directions. 
This induces a directional gradient dominance at the first optimization step, preventing convergence to symmetric equilibria. 
Empirically, moderate asymmetry (e.g., 0.7/0.3) is sufficient.

\subsection{Final DAG Check}
\label{sec:final_dag_check}

After training, we perform a post-hoc DAG check on the extracted variable-level graph.
From the learned state-level adjacency $\mathbf{A}$, we extract directed edges by block-max thresholding:
\[
\mathcal{E}_{\mathrm{raw}}
=
\left\{
(i,j)\; \middle|\;
\max_{a\in\mathcal I_i,\; b\in\mathcal I_j} A_{ab} > \tau,\; i\neq j
\right\}.
\]

We then test acyclicity on $\mathcal{E}_{\mathrm{raw}}$.
If cycle projection is enabled, each detected directed cycle is broken by removing the \emph{weakest} edge, where edge strength is defined as the mean value of its state block:
\[
\operatorname{str}(i\!\to\!j)
=
\frac{1}{|\mathcal I_i||\mathcal I_j|}
\sum_{a\in\mathcal I_i}\sum_{b\in\mathcal I_j} A_{ab}.
\]
The edge with the smallest $\operatorname{str}(i\!\to\!j)$ on that cycle is cut, and this is repeated until no cycle remains. Note that this step is post-training and does not alter the optimization objective. It ensures that the final reported graph is explicitly checked (and, if enabled, projected) for DAG validity. Algo.~\ref{alg:refinement} outlines our overall approach.

\section{Experimental Evaluation}
CausalSAGE is specifically designed to resolve orientation ambiguity in partially directed causal graphs, where constraint-based discovery methods(e.g., FCI/RFCI) often leave a large fraction of edges unoriented due to Markov equivalence.
Accordingly, our experiments focus on evaluating whether such post-discovery refinement can reliably transform a PAG into a fully directed DAG while improving structural accuracy and maintaining scalability. Unless otherwise specified, all hyperparameters are fixed across datasets (see Appendix~\ref{sec:hyper_values}).

Overall,we aim to answer the following questions:

\textbf{(Q1) Refinement effectiveness:} Can our framework reliably refine a PAG estimated by a constraint-based method into a fully directed DAG?
\textbf{(Q2) Comparison to direct DAG learners:} How does our approach compare with standard DAG learners that operate directly on fully observed data?
\textbf{(Q3) Sensitivity to Sample Size and Random Initialization:} How does refinement performance vary with observational sample size and random initialization, and to what extent does the method depend on specific data regimes or initialization choices?
\textbf{(Q4) Scalability:} How does runtime scale with graph size, and can our method handle hundreds of variables efficiently?

\paragraph{Datasets.}
We evaluate our method on standard benchmark Bayesian networks obtained from the \texttt{bnlearn} repository \citep{scutari2010learning}. These networks are widely used in causal structure learning and provide ground-truth DAGs for evaluation.Unless otherwise specified, we use a fixed sample size per dataset. Details of benchmark datasets are provided in Appendix~\ref{sec:dataset_details}.

The benchmarks span small to large graph sizes, ranging from 11 variables (Sachs) to 724 variables (Link) which allow the assessment of both the structural and scalability aspects of our approach.
For each benchmark, we generate observational samples from the ground-truth DAG and apply identical preprocessing across all methods. 

Although all benchmarks are fully observed, we employ FCI(RFCI) as a generic constraint-based method to obtain an initial partially directed structure. Our refinement framework then aims to break the Markov equivalence and recover a fully directed DAG.

\paragraph{Evaluation Metrics.}
We evaluate structural accuracy using Structural Hamming Distance (SHD) and F1 score between the estimated DAG and the ground-truth DAG.
To quantify orientation ambiguity, we additionally report the \emph{unresolved direction ratio}, defined as the proportion of skeleton edges for which both directions receive high confidence after optimization.
Lower values indicate fewer symmetric ambiguities.
Runtime is reported to assess computational efficiency.

\subsection{Refinement over PAG (Q1)}
We evaluate whether our refinement framework can transform a partially oriented PAG, estimated by FCI/RFCI, into an accurate fully directed DAG.
We compare against (i) raw FCI/RFCI outputs, which retain orientation ambiguity, and (ii) FCI augmented with LLM-based orientation suggestions (\textsc{FCI+LLM}). 
This comparison isolates the effect of structural refinement from prior injection and allows us to examine whether the improvement is due to the refinement mechanism, rather than simply injecting prior information.

\paragraph{Results.}

\begin{figure}[t]
    \centering
    \includegraphics[width=\linewidth]{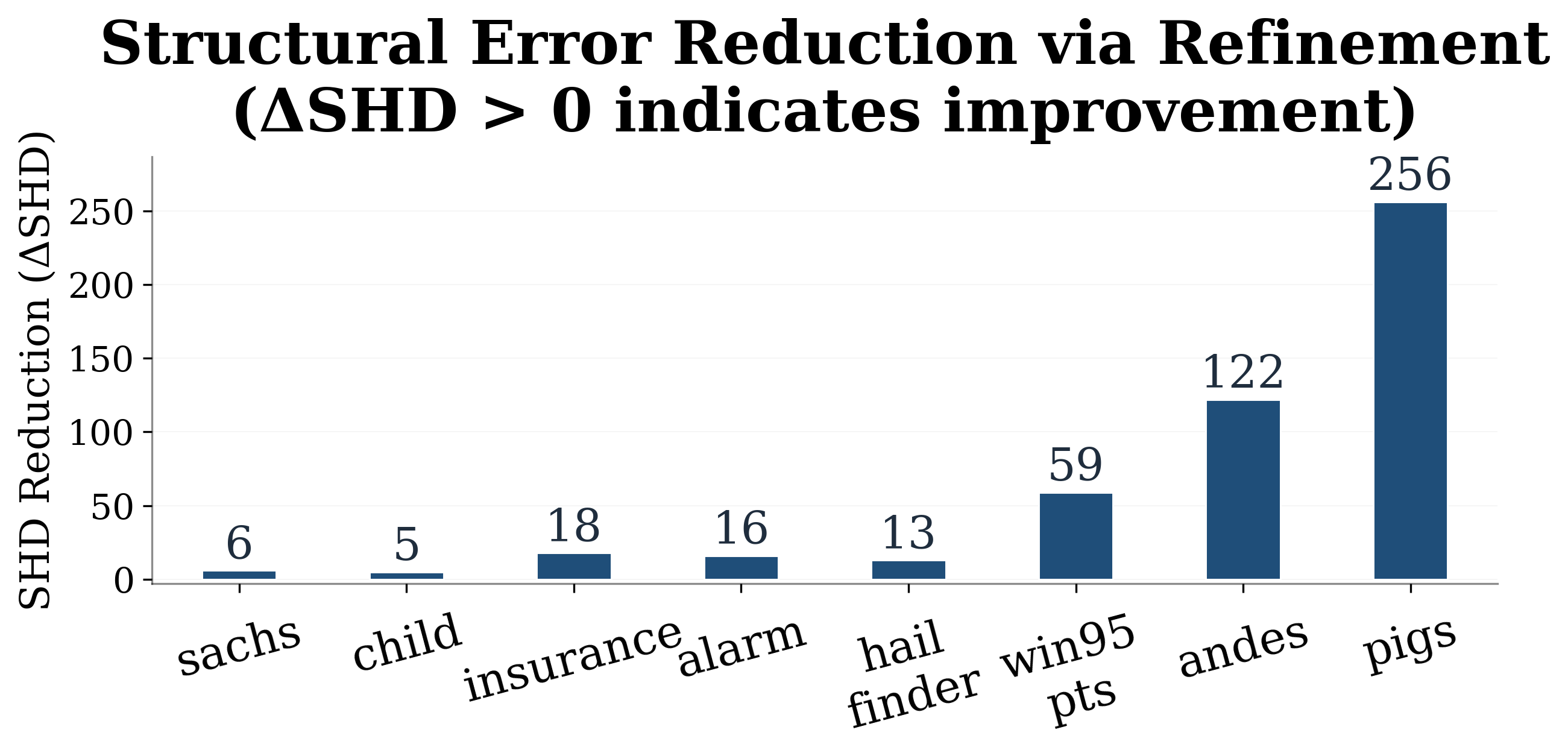}
    \caption{
    \textbf{SHD reduction via refinement (Q1).}
    Bars show $\Delta$SHD = (Baseline $-$ Refined), where the positive values indicate improvement.}
    \label{fig:shd_reduction}
\end{figure}

Figure~\ref{fig:shd_reduction} reports $\Delta$SHD across datasets. 
Refinement consistently reduces structural error compared to raw FCI/RFCI, with especially large improvements on medium and large graphs. 
For example, on \textit{andes}, SHD decreases from 271 to 149 ($\Delta=122$), and on \textit{pigs}, from 276 to 20 ($\Delta=256$). 
FCI+LLM yields intermediate improvements but remains substantially worse than our refinement (full results in Appendix ~\ref{sec:effectiveness}).

\begin{figure}[t]
    \centering
    \includegraphics[width=\linewidth]{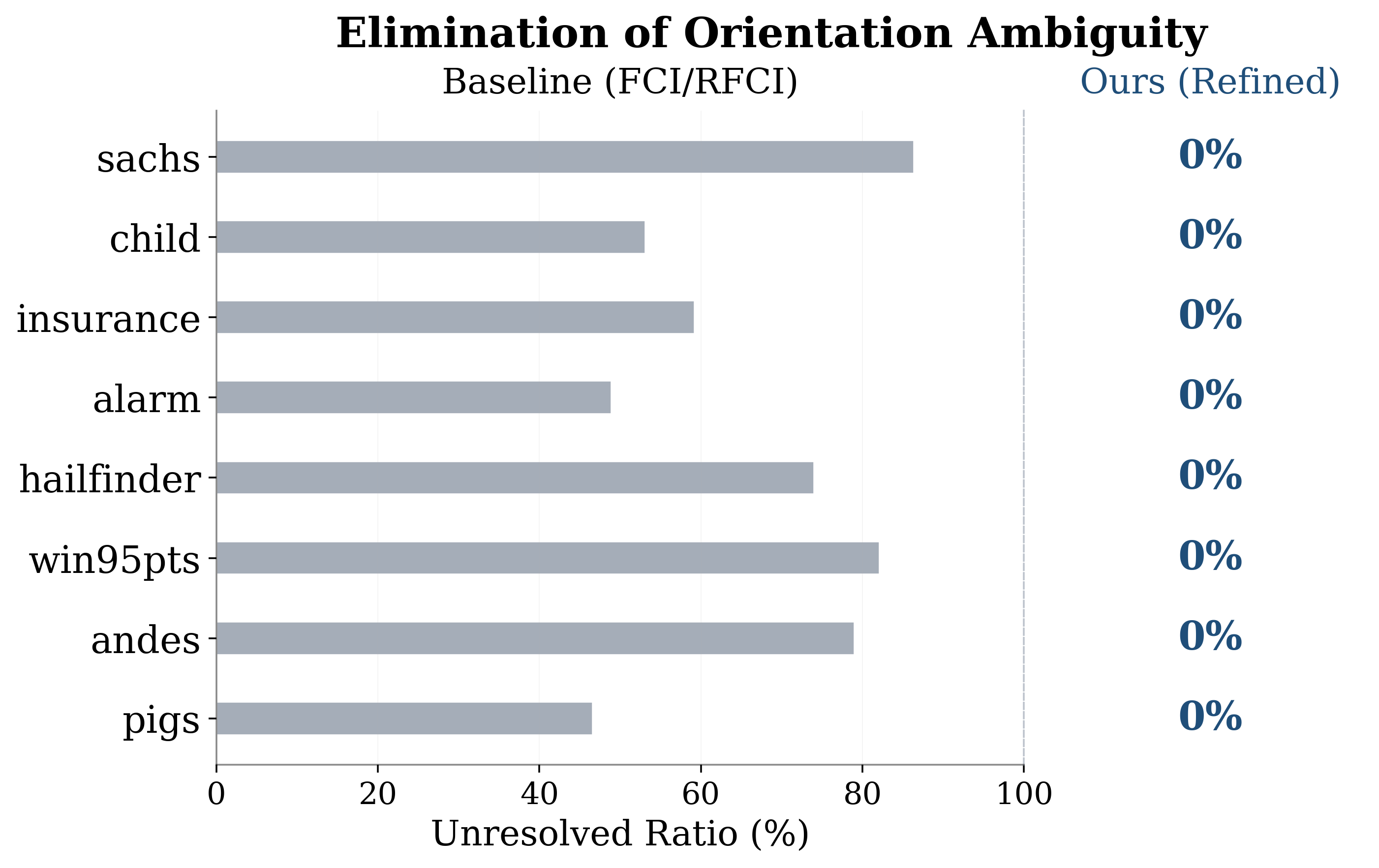}
    \caption{\textbf{Elimination of Orientation Ambiguity.}
    FCI/RFCI leaves a large fraction of edges unresolved (46\%–86\%) across datasets. CausalSAGE reduces this ratio to 0\% in all cases, effectively breaking Markov equivalence and producing fully directed DAGs.}
    \label{fig:unresolved_rito}
\end{figure}

In addition, the unresolved direction ratio (the definition is in Appendix~\ref{sec:definition}.) drops from 46–86\% under FCI/RFCI to 0\% after refinement across all benchmarks (Table~\ref{fig:unresolved_rito}). 
In a small number of cases under the LLM prior, a residual fraction of ambiguous edges is resolved during the final DAG validation step, ensuring complete orientation.

\paragraph{DAG Validity.} In most cases, optimization directly converges to a valid DAG without post-processing (6 out of 9 datasets under random prior). 
When directed cycles occur, a lightweight correction step removes only a small number of edges (typically 1–6 per dataset), accounting for less than 3\% of total edges even on large graphs such as \textit{andes}. 
This indicates that the refinement objective converges to near-acyclic solutions with minimal adjustment. Details of DAG Validity evaluation are in ~\ref{sec:DAG_check}.
These results affirmatively answer (Q1): CausalSAGE reliably transforms partially oriented PAGs into accurate, fully directed causal graphs.

\subsection{Comparison with DAG learners (Q2)}
Since CausalSAGE outputs a fully directed DAG, we additionally compare it against standard DAG learners.

\paragraph{Baselines.} 
(i) PC, a constraint-based method relying on conditional independence tests;
(ii) MMHC, a hybrid approach combining constraint-based skeleton learning with score-based search;
(iii) Tabu search and (iv) Hill Climbing (HC), two score-based optimization methods commonly used in Bayesian network learning.

\begin{figure*}[t]
    \centering
    \includegraphics[width=\linewidth]{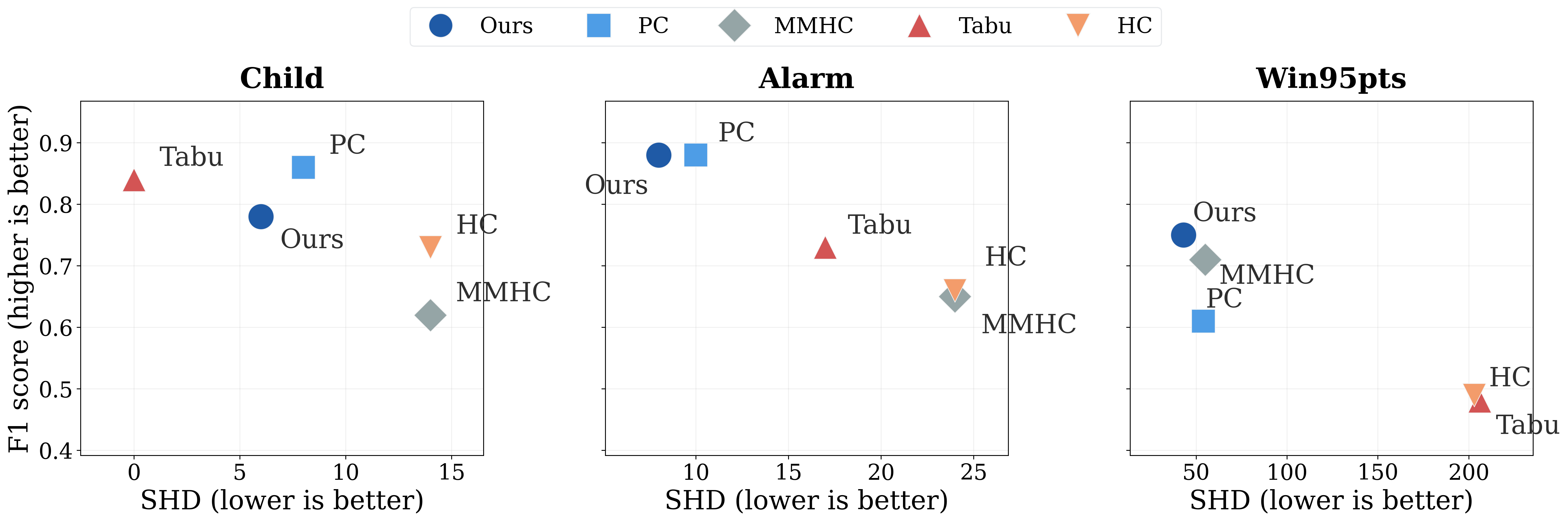}
    \caption{
\textbf{Comparison with standard DAG learners (10k samples).}
SHD measures structural error,
while F1 measures directional accuracy.
Points closer to the top-left corner indicate stronger overall performance.}
    \label{fig:shd_f1_pareto_scatter}
\end{figure*}
\paragraph{Results.}
We compare our refinement approach with classical direct DAG learners across all fully observed discrete benchmarks. Due to space constraints, we report three representative datasets in Figure~\ref{fig:shd_f1_pareto_scatter}, with complete results provided in the Appendix ~\ref{full comparison}.
Points closer to the top-left corner (lower SHD, higher F1) indicate stronger structural recovery.
Across datasets, our method consistently achieves competitive performance and remains near the Pareto frontier.
Importantly, unlike certain score-based methods that exhibit large structural degradation on medium-scale graphs (e.g., Tabu on \textit{win95pts}), our refinement approach does not suffer from catastrophic failures as graph size increases.


Although some score-based methods achieve strong performance on smaller networks (e.g., Tabu on \textit{child}), our approach demonstrates greater stability across datasets and better scalability as graph size increases.
In addition, our refinement framework scales to substantially larger networks, 
including the \textit{link} dataset with 724 variables. 
In contrast, several classical score-based DAG learners become computationally expensive or unstable on very large graphs, making direct comparison impractical at this scale under our experimental setup. 
Overall, we can answer (Q2) affirmatively: the proposed refinement strategy provides a robust and balanced alternative to classical direct DAG learners.

\subsection{Sensitivity to Sample Size and Random Seeds (Q3)}
We examine how the refinement performance varies with the number of observational samples and random initialization.
We vary the sample size: $N \in \{1\text{k}, 2\text{k}, 5\text{k}, 10\text{k}\} $ and evaluate robustness across 5 random seeds $\text{seed} \in \{1,2,3,4,5\}$ while keeping all hyperparameters fixed.
\begin{figure}[t]
    \centering
    \includegraphics[width=\linewidth]{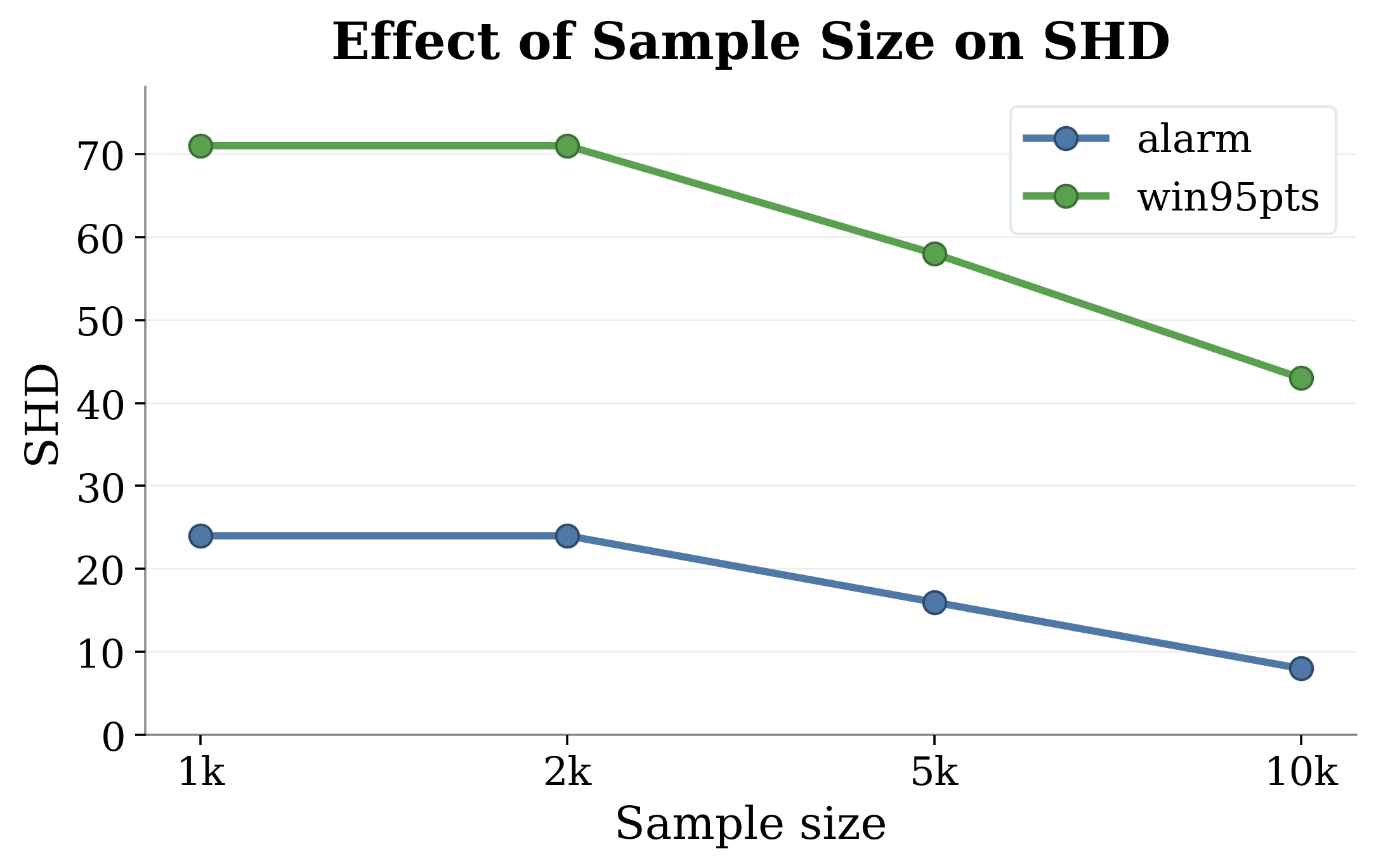}
    \caption{
\textbf{Effect of sample size on SHD.}
As the number of observational samples increases, structural error consistently decreases across datasets.}
    \label{fig:sample_size}
\end{figure}
\paragraph{Results.}
Figure~\ref{fig:sample_size} shows that for medium and large graphs (e.g., \textit{alarm} and \textit{win95pts}), SHD generally decreases as $N$ increases, reflecting improved skeleton estimation and subsequent refinement.
Smaller networks (e.g., \textit{child}) exhibit mild fluctuations, indicating higher variance under limited samples.
Overall, performance improves with more data but does not critically depend on a specific sample regime. Detailed results for all datasets are provided in the Appendix ~\ref{sample_size}.
Moreover, most of the variation across sample sizes can be attributed to changes in the quality of the FCI/RFCI skeleton.
When the initial skeleton is fixed, refinement results remain largely stable across different sample regimes.

\paragraph{Random seeds.}
Across multiple random seeds, refinement performance is generally stable on small and medium-sized graphs. On larger graphs, the random prior exhibits moderate variance, while the llm-based prior consistently shows near-zero variance. This indicates that semantic initialization improves optimization consistency, particularly for larger networks.
Complete results in Appendix ~\ref{sec:random_seeds}.

\subsection{Scalability (Q4).}
We measure runtime of CausalSAGE as a function of graph size.
For large datasets (\textit{pigs}, \textit{link}), we employ mini-batch optimization with a batch size of 2048 to reduce memory consumption and stabilize gradient updates.
All experiments were on a single Intel Core Ultra 7 CPU with 32GB RAM.
\begin{table}[t]
\centering
\small
\caption{
\textbf{CausalSAGE is efficient.} The runtime as a function of graph size shows the efficiency of our method.
}
\begin{tabular}{lrrr}
\toprule
\textbf{Dataset} & \textbf{Variables} & \textbf{Sample Size} & \textbf{Runtime (s)} \\
\midrule
Sachs      & 11  & 855   & 2.8  \\
Child      & 20  & 5000  & 2.6  \\
Insurance  & 27  & 10000 & 38   \\
Alarm      & 37  & 10000 & 32   \\
Hailfinder & 56  & 10000 & 59   \\
Win95pts   & 76  & 10000 & 71   \\
Andes      & 223 & 10000 & 189  \\
Pigs       & 441 & 50000 & 368  \\
Link       & 724 & 50000 & 742  \\
\bottomrule
\end{tabular}
\label{tab:runtime}
\end{table}

Table~\ref{tab:runtime} reports the runtime growth across datasets.
Runtime grows approximately linearly with the number of variables in practice,
ranging from a few seconds for small graphs (e.g., \textit{child}, \textit{sachs})
to approximately 3 minutes for \textit{andes} (223 variables),
6 minutes for \textit{pigs} (441 variables),
and 12 minutes for the largest benchmark \textit{link} (724 variables). A detailed computational complexity analysis is provided in Appendix~\ref{sec:computational complexity}.
Importantly, CausalSAGE remains feasible even at the scale of several hundred variables, successfully handling the 724-node \textit{link} network on a single CPU. Complete runtime of FCI/RFCI is in Appendix~\ref{sec:runtime}.
These results answer (Q4) affirmatively: CausalSAGE scales to large causal graphs without prohibitive computational costs.

\section{Conclusion}
We present CausalSAGE, a refinement framework for converting PAGs learned from standard constraint-based causal discovery methods to DAGs. We demonstrate that CasualSAGE can efficiently perform the conversion based on a unified objective function that optimizes directional preference. After training, the framework also performs a post-hoc DAG check in order to ensure acyclicity.

Future work involves testing CausalSAGE on real world applications such as molecular networks~\citep{kelly2022review,brown2025large} and climate change~\citep{ebert2012causal,ombadi2020evaluation}. Extending CausalSAGE to incorporate dynamical systems is an interesting future direction. Finally, constructing more informed priors is an immediate next step.


\bibliography{references}

\newpage
\onecolumn
\title{No More Maybe-Arrows: \\Resolving Causal Uncertainty by Breaking Symmetries \\(Supplementary Material)}
\maketitle
\appendix

\section{Additional Theoretical Details}

\subsection{Gradient Symmetry and Interior Equilibrium}
\label{app:symmetry_analysis}

We analyze the gradient dynamics for one unresolved variable pair $(i,j)$
, for which both directions are admissible.

\paragraph{Block magnitudes.}
Let
\[
\alpha = \|\mathbf{A}_{ij}\|_F,
\qquad
\beta  = \|\mathbf{A}_{ji}\|_F,
\]
denote the directional block magnitudes,
where $\mathbf{A}=\sigma(\mathbf{W})\odot\mathbf{S}$.

The pairwise cycle penalty contributes
\[
\mathcal{L}_{\text{cycle}}^{(i,j)} = \alpha\beta .
\]

\paragraph{Gradient decomposition.}
Restricting the objective to this pair,
\[
\mathcal{L}_{\text{pair}}
=
\mathcal{L}_{\text{recon}}
+
\lambda_2 \alpha\beta .
\]

The partial derivatives with respect to $\alpha$ and $\beta$ are
\begin{equation}
\frac{\partial \mathcal{L}_{\text{pair}}}{\partial \alpha}
=
g_\alpha(\alpha,\beta)
+
\lambda_2 \beta,
\label{eq:dalpha_app}
\end{equation}
\begin{equation}
\frac{\partial \mathcal{L}_{\text{pair}}}{\partial \beta}
=
g_\beta(\alpha,\beta)
+
\lambda_2 \alpha,
\label{eq:dbeta_app}
\end{equation}
where
\[
g_\alpha = \frac{\partial \mathcal{L}_{\text{recon}}}{\partial \alpha},
\qquad
g_\beta  = \frac{\partial \mathcal{L}_{\text{recon}}}{\partial \beta}.
\]

\paragraph{Interior stationary point.}
A stationary point $(\alpha^\star,\beta^\star)$ satisfies
\begin{equation}
g_\alpha(\alpha^\star,\beta^\star)
+
\lambda_2 \beta^\star
=
0,
\qquad
g_\beta(\alpha^\star,\beta^\star)
+
\lambda_2 \alpha^\star
=
0.
\label{eq:equilibrium_app}
\end{equation}

If $g_\alpha$ and $g_\beta$ have comparable magnitudes near symmetry
(e.g., under finite-sample uncertainty or MEC ambiguity),
system \eqref{eq:equilibrium_app} admits an interior solution
with
\[
\alpha^\star > 0,
\qquad
\beta^\star  > 0.
\]

\paragraph{Local curvature analysis.}
The Hessian of $\mathcal{L}_{\text{pair}}$ with respect to $(\alpha,\beta)$ is
\[
\mathbf{H}
=
\begin{pmatrix}
\frac{\partial^2 \mathcal{L}_{\text{recon}}}{\partial \alpha^2}
&
\frac{\partial^2 \mathcal{L}_{\text{recon}}}{\partial \alpha\partial \beta} + \lambda_2
\\
\frac{\partial^2 \mathcal{L}_{\text{recon}}}{\partial \beta\partial \alpha} + \lambda_2
&
\frac{\partial^2 \mathcal{L}_{\text{recon}}}{\partial \beta^2}
\end{pmatrix}.
\]

Under a local regime where second-order reconstruction terms are small
and cross-couplings are weak,
the off-diagonal terms are dominated by $\lambda_2$.
In this case,
\[
\det(\mathbf{H})
\approx
-\lambda_2^2 < 0,
\]
indicating that the interior stationary point exhibits saddle-type behavior
along asymmetric directions.

\paragraph{Linear approximation.}
Under a local approximation
\[
g_\alpha \approx -c_\alpha,
\qquad
g_\beta  \approx -c_\beta,
\quad
c_\alpha, c_\beta > 0,
\]
the stationary point satisfies
\[
\beta^\star = \frac{c_\alpha}{\lambda_2},
\qquad
\alpha^\star = \frac{c_\beta}{\lambda_2}.
\]

Hence both directions remain nonzero unless one of
$c_\alpha, c_\beta$ vanishes.

\paragraph{Effect of asymmetric initialization.}
If initialization satisfies $\alpha_0 \gg \beta_0$,
then from \eqref{eq:dbeta_app},
\[
\frac{\partial \mathcal{L}_{\text{pair}}}{\partial \beta}
\approx
g_\beta + \lambda_2 \alpha_0,
\]
which can become positive even when $g_\beta < 0$,
driving $\beta$ toward zero and eliminating the interior equilibrium.

\subsection{Derivation of the Cycle Regularizer}
\label{sec:cycle_analysis}

For each unresolved variable pair $(i,j)$,
we define directional block magnitudes
\[
\alpha = \|\mathbf{A}_{ij}\|_F,
\qquad
\beta  = \|\mathbf{A}_{ji}\|_F.
\]

\paragraph{Design objective.}
The regularizer is intended to satisfy:

\begin{enumerate}
    \item No penalty if either direction is absent:
    \[
    \alpha = 0 \quad \text{or} \quad \beta = 0
    \;\Rightarrow\;
    \mathcal{L}_{\text{cycle}} = 0.
    \]
    
    \item Increasing penalty when both directions are simultaneously active.
    
    \item Differentiability for gradient-based optimization.
\end{enumerate}

The simplest smooth function satisfying these properties is the bilinear coupling
\[
\mathcal{L}_{\text{cycle}}^{(i,j)} = \alpha\beta.
\]

\paragraph{Gradient structure.}
The partial derivatives are
\[
\frac{\partial}{\partial \alpha} (\alpha\beta) = \beta,
\qquad
\frac{\partial}{\partial \beta} (\alpha\beta) = \alpha.
\]

Thus, growth in one direction linearly increases the penalty gradient on the opposite direction,
inducing mutual suppression.

\paragraph{Comparison to alternative penalties.}
An additive penalty $\alpha + \beta$ would penalize single-direction activation
and bias both directions toward zero.
A quadratic penalty $\alpha^2 + \beta^2$ similarly shrinks each direction independently.
In contrast, the bilinear term $\alpha\beta$ penalizes only co-activation,
leaving single-direction solutions unpenalized.
\subsection{Computational Complexity Analysis}
\label{sec:computational complexity}
We analyze the computational and memory complexity of the proposed refinement procedure under the current dense implementation.

\paragraph{Notation.}
Let $M$ denote the number of variables and $L_i$ the number of states of variable $V_i$.
The total state dimension is
\[
n_s = \sum_{i=1}^{M} L_i.
\]
Let $N$ denote the number of samples and $B$ the mini-batch size.
The state-level mask is $\mathbf{S}\in\{0,1\}^{n_s\times n_s}$.
All matrix operations are implemented using dense tensors.

\paragraph{Forward pass.}
In each optimization step, masked logits are computed as
\[
\mathbf{L}
=
\mathbf{X}\left(\mathbf{W}\odot\mathbf{S}\right),
\quad
\mathbf{X}\in\{0,1\}^{B\times n_s},
\quad
\mathbf{W}\in\mathbb{R}^{n_s\times n_s}.
\]

The computation consists of two dense operations:

\begin{enumerate}
    \item Element-wise masking:
    \[
    \mathbf{W}_{\text{masked}} = \mathbf{W}\odot\mathbf{S},
    \]
    which requires $\mathcal{O}(n_s^2)$ operations.
    
    \item Dense matrix multiplication:
    \[
    \mathbf{L} = \mathbf{X}\mathbf{W}_{\text{masked}},
    \]
    which requires
    \[
    \mathcal{O}(B\,n_s^2)
    \]
    operations.
\end{enumerate}

The per-variable softmax and cross-entropy computation over the expanded states requires
\[
\mathcal{O}(B\,n_s),
\]
which is dominated by the matrix multiplication term.

Therefore, the total forward-pass complexity per optimization step is
\[
\boxed{
\mathcal{O}(B\,n_s^2)
}.
\]

\paragraph{Backward pass.}
Backpropagation through the dense matrix multiplication has the same asymptotic complexity as the forward pass, yielding
\[
\mathcal{O}(B\,n_s^2).
\]

Thus, each gradient step requires
\[
\boxed{
\mathcal{O}(B\,n_s^2)
}.
\]

\paragraph{Regularization terms.}
All structural regularizers ($\mathcal{L}_{\text{sparse}}$, 
$\mathcal{L}_{\text{cycle}}$, and 
$\mathcal{L}_{\text{skeleton}}$)
operate on the masked adjacency matrix
$\mathbf{A}=\sigma(\mathbf{W})\odot\mathbf{S}$.
Since these terms involve Frobenius norms or element-wise operations over
$n_s\times n_s$ matrices,
their cost is bounded by
\[
\mathcal{O}(n_s^2),
\]
which is dominated by the reconstruction term when $B \ge 1$.

\paragraph{Post-processing.}
After training, block-wise thresholding requires scanning all state-level blocks, which costs $\mathcal{O}(n_s^2)$.
Cycle detection on the extracted variable-level graph costs
$\mathcal{O}(M + |\mathcal{E}|)$,
where $|\mathcal{E}|$ is the number of extracted edges.
Since $M \ll n_s$ in typical multi-state settings,
post-processing is negligible compared to optimization.

\paragraph{Space complexity.}
The dominant memory cost arises from storing the dense parameter matrix
$\mathbf{W}\in\mathbb{R}^{n_s\times n_s}$,
which requires
\[
\boxed{
\mathcal{O}(n_s^2)
}
\]
memory.
Mini-batch data storage requires $\mathcal{O}(B\,n_s)$.

\paragraph{Discussion.}
In the worst case, the refinement procedure scales quadratically in the total state dimension $n_s$ per optimization step.
Since $n_s=\sum_i L_i$, this corresponds to
$\mathcal{O}(B\,(\sum_i L_i)^2)$.

Importantly, this quadratic dependence is with respect to the \emph{state dimension}, not directly the number of variables.
In many standard benchmark networks, variables have small cardinalities (typically binary or ternary),
so $n_s \approx cM$ with a small constant $c$.
For example, in the largest benchmark (\textit{link}, 724 variables),
$n_s$ remains on the order of $10^3$, resulting in a dense parameter matrix of manageable size.
Combined with highly optimized BLAS matrix multiplication,
this yields practical runtimes of under 12 minutes on a single CPU,
as reported in Sec.~4.4.

Nevertheless, the quadratic dependence suggests that the current dense implementation is best suited to low- and moderate-cardinality discrete variables.
Extending the framework to high-cardinality settings would benefit from sparse or block-structured implementations, which we leave for future work.

\section{Implementation Details}
\label{sec:appendix_impl}

\subsection{Dataset Details}
\label{sec:dataset_details}

\begin{table}[htp]
\centering
\caption{Benchmark datasets from the bnlearn repository.}
\label{tab:datasets}
\begin{tabular}{lcccc}
\toprule
Dataset & Nodes & Edges & Samples & Type \\
\midrule
Sachs & 11 & 17 & 855 & Continuous \\
Child & 20 & 25 & 5000 & Discrete \\
Insurance & 27 & 52 & 10000 & Discrete \\
Alarm & 37 & 46 & 10000 & Discrete \\
Hailfinder & 56 & 66 & 10000 & Discrete \\
Win95pts & 76 & 112 & 10000 & Discrete \\
Andes & 223 & 338 & 10000 & Discrete \\
Pigs & 441 & 592 & 50000 & Discrete \\
Link & 724 & 1125 & 50000 & Discrete \\
\bottomrule
\end{tabular}
\end{table}

\subsection{Definition of Unresolved direction Ratio}
\label{sec:definition}
Let $\mathcal Q$ denote the set of unordered variable pairs
for which both directions are structurally admissible
under the input PAG (i.e., unresolved pairs).

For each $\{i,j\} \in \mathcal Q$,
we define the directional strength using the block-wise aggregation
introduced in Sec.~\ref{sec:dir_softmax}:
\[
s_{i\to j}
=
\max_{a\in \mathcal I_i,\; b\in \mathcal I_j}
A_{ab},
\qquad
s_{j\to i}
=
\max_{a\in \mathcal I_j,\; b\in \mathcal I_i}
A_{ab},
\]
where $\mathbf A = \sigma(\mathbf W)\odot\mathbf S$
is the learned state-level adjacency matrix.

Given a threshold $\tau$,
a pair $\{i,j\}$ is considered \emph{unresolved}
if both directions exceed the threshold:
\[
\{i,j\} \text{ is unresolved}
\iff
s_{i\to j} > \tau
\;\land\;
s_{j\to i} > \tau.
\]

The neural unresolved ratio is defined as
\[
r_{\mathrm{neural}}
=
\frac{
\sum_{\{i,j\}\in\mathcal Q}
\mathbf I
\bigl(
s_{i\to j} > \tau
\land
s_{j\to i} > \tau
\bigr)
}{
|\mathcal Q|
},
\]
where $\mathbf I(\cdot)$ denotes the indicator function.

\subsection{Optimization Setup}
\label{sec:setup_details}

We optimize the unified objective in Eq.~(8) using Adam with learning rate $\eta = 0.01$
for $T = 140$ optimization steps.
All hyperparameters are fixed across datasets.

For small and medium-sized graphs, full-batch optimization is used.
For large graphs (\textit{pigs}, \textit{link}), we employ mini-batch training with batch size 2048 to reduce memory consumption and stabilize gradient updates.

No early stopping or dataset-specific tuning is applied.
The diagonal entries of $W$ are fixed to $-\infty$ to prevent self-loops.
State-level normalization ensures that the optimization scale remains comparable across different graph sizes.

\subsection{Hyperparameter Values}
\label{sec:hyper_values}
Unless otherwise specified, we use
$\lambda_1 = 0.01$,
$\lambda_2 = 5$,
$\lambda_3 = 0.1$,
and extraction threshold $\tau = 0.1$.

All hyperparameters are fixed across datasets and random seeds.

\subsection{Prior Initialization}
\label{prior_details}

Initialization consists of two components: structural masking and directional bias.

First, the input PAG is converted into a state-level hard mask $\mathbf{S}$,
which determines the admissible state-to-state connections.
Entries with $S_{ab}=0$ remain fixed to zero throughout training.

Second, for each unresolved variable pair $(V_i,V_j)$,
we assign an asymmetric initialization to the corresponding state-level blocks.
The favored direction is initialized with probability $0.9$,
and the opposite direction with probability $0.1$.
These probabilities are converted to logits to initialize $\mathbf{W}$.

Under the \textbf{random prior}, the favored direction is sampled uniformly at random.

Under the \textbf{LLM-based prior}, GPT-3.5-turbo is queried using variable names to estimate directional plausibility.
The predicted preference is mapped to the same $0.9/0.1$ initialization scheme.

Importantly, the prior only affects initialization.
It does not modify the structural mask or the optimization objective,
and both directions remain admissible during training.

\section{Complete results and tables}

Direct DAG learners comparasion, sample size, Random Seeds Variance, Runtime Full Table(give FCI runtime)

\subsection{Refinement Effectiveness (SHD)}
\label{sec:effectiveness}

Table~\ref{tab:shd_results} reports the Structural Hamming Distance (SHD) for all benchmark datasets under a fixed random seed (seed = 5). 
We compare the raw FCI/RFCI outputs, FCI augmented with LLM-based orientation suggestions (FCI+LLM), and our refinement method under both random and LLM-based priors.

\begin{table}[t] 
    \centering
    \caption{Structural Hamming Distance (SHD) Comparison(random seed=5)}
    \label{tab:shd_results}
    \begin{tabular}{lrrrrr}
        \toprule
        Dataset & Variables & FCI(RFCI) & FCI + LLM & Ours (LLM Prior) & Ours (Random Prior) \\
        \midrule
        sachs      & 11 & 14  & 10  & 9  & 8  \\
        child      & 20 & 11  & 7   & 7   & 6   \\
        insurance  & 27 & 32  & 23  & 21 & 13  \\
        alarm      & 37 & 24  & 12  & 11  & 8  \\
        hailfinder & 56 & 58  & 42  & 46 & 45  \\
        win95pts   & 76 & 102  & 46  & 44 & 43  \\
        andes      & 223 & 271 & 198 & 149 & 152 \\
        pigs       & 441 & 276 & 270 & 20  & 20   \\
        link       & 724 & 947 & 834 &  663   & 645 \\
        \bottomrule
    \end{tabular}
\end{table}

\subsection{Comparison vs DAG learners (SHD + F1)}
\label{full comparison}
Since FCI/RFCI outputs partially oriented graphs, directed edge F1 is only reported for methods that produce fully directed DAGs. The F1 results are shown in Table~\ref{tab:full_f1}, and the SHD results are shown in Table~\ref{tab:full_shd_dag}

\begin{table}[t]
\centering
\small
\setlength{\tabcolsep}{6pt}
\caption{Directed edge F1 comparison for methods that output fully directed DAGs. Higher is better.}
\begin{tabular}{l r rrrrrr}
\toprule
Dataset & Variables & PC & MMHC & Tabu & HC & Ours (Random) & Ours (LLM) \\
\midrule
Child      & 20  & 0.86 & 0.62 & 0.84 & 0.73 & 0.77 & 0.78 \\
Insurance  & 27  & 0.62 & 0.5 & 0.66 & 0.60 & 0.70 & 0.83 \\
Alarm      & 37  & 0.88 & 0.65 & 0.73 & 0.66 & 0.80 & 0.88 \\
Hailfinder & 56  & 0.54 & 0.67 & 0.81 & 0.86 & 0.53 & 0.53 \\
Win95pts   & 76  & 0.61 & 0.71 & 0.48 & 0.49 & 0.72 & 0.75 \\
Andes      & 223 & 0.68 & 0.87 & 0.75 & 0.75 & 0.61 & 0.65 \\
Pigs       & 441 & 0.79 & 0.79 & 0.97 & 0.83 & 0.98 & 0.98 \\
Link       & 724 & - & 0.15 & - & - & 0.57 & 0.47 \\
        \bottomrule
\end{tabular}
\label{tab:full_f1}
\end{table}
\begin{table}[t]
\centering
\small
\caption{Structural Hamming Distance (SHD) comparison with direct DAG learners. (Seeds=5).}
\setlength{\tabcolsep}{6pt}
\begin{tabular}{l r rrrrrr}
\toprule
Dataset & Variables & PC & MMHC & Tabu & HC & Ours (Random) & Ours (LLM) \\
\midrule
Child      & 20  & 8 & 14 & 0 & 14 & 6 & 7 \\
Insurance  & 27  & 33 & 36 & 36 & 43 & 13 & 21 \\
Alarm      & 37  & 10 & 24 & 17 & 24 & 8 & 11 \\
Hailfinder & 56  & 47 & 33 & 36 & 20 & 45 & 46 \\
Win95pts   & 76  & 54 & 55 & 206 & 203 & 43 & 44\\
Andes      & 223 & 152 & 85 & 229 & 227 & 152 & 149 \\
Pigs       & 441 & 218 & 193 & 10 & 163 & 20 & 20 \\
Link       & 724 & - & 918 & - & - & 645 & 663 \\
        \bottomrule
\end{tabular}
\label{tab:full_shd_dag}
\end{table}

\subsection{Sensitivity to Sample Size}
\label{sample_size}
Table~\ref{tab:sample_sizes} reports the SHD obtained under varying 
observational sample sizes ($N \in \{1k, 2k, 5k, 10k\}$), while keeping all hyperparameters fixed. These results provide the complete numerical values corresponding to Fig.~\ref{fig:sample_size} in the main text.
to Fig.~\ref{fig:sample_size} in the main text.
\begin{table}[t]
\centering
\small
\caption{SHD under varying sample sizes.}
\label{tab:sample_sizes}
\begin{tabular}{l rrrrr}
\toprule
Dataset & 1k & 2k & 5k & 10k \\
\midrule
Sachs      & 13 & 14 & 32 & 9 \\
Child      & 12 & 4 & 5 & 6 \\
Insurance  & 25 & 24 & 22 & 13 \\
Alarm      & 24 & 24 & 16 & 8 \\
Hailfinder & 40 & 41 & 37 & 44 \\
Win95pts   & 71 & 71 & 58 & 43 \\
Andes      & 226 & 194 & 147 & 151 \\
\bottomrule
\end{tabular}
\end{table}

\paragraph{Fixed skeleton analysis.}
To isolate the effect of sample size on the refinement procedure itself,
we fix the input PAG skeleton (estimated at $N=10\text{k}$)
and vary only the observational samples used in the reconstruction objective. Table~\ref{tab:fixed_skeletons} reports the corresponding SHD results.

\begin{table}[t]
\centering
\small
\caption{SHD under fixed skeletons.}
\label{tab:fixed_skeletons}
\begin{tabular}{l rrrrr}
\toprule
Dataset & 1k & 2k & 5k & 10k \\
\midrule
Sachs      & 9 & 9 & 9 & 9 \\
Child      & 6 & 6 & 6 & 6 \\
Insurance  & 13 & 13 & 13 & 13 \\
Alarm      & 8 & 8 & 8 & 8 \\
Hailfinder & 41 & 41 & 41 & 44 \\
Win95pts   & 43 & 43 & 43 & 43 \\
Andes      & 151 & 151 & 151 & 151 \\
\bottomrule
\end{tabular}
\end{table}
With the skeleton fixed, SHD remains largely stable across different sample sizes, indicating that most variation observed in Fig.~\ref{fig:sample_size} arises from differences in the estimated PAG structure rather than from the refinement mechanism itself.

\subsection{Stability Across Random Seeds}
\label{sec:random_seeds}

Table~\ref{random_seeds} reports the mean and standard deviation 
of SHD over five independent random seeds, with all hyperparameters fixed.

\begin{table}[t]
\centering
\small
\caption{SHD mean $\pm$ std over 5 random seeds.}
\begin{tabular}{l rr}
\toprule
Dataset & Ours (Random) & Ours (LLM) \\
\midrule
Sachs      & 8 $\pm$ 4.8      & 9 $\pm$ 0.0 \\ 
Child      & 5.4 $\pm$ 2.64   & 7 $\pm$ 0.0 \\
Insurance  & 17.4 $\pm$ 4.24  & 21 $\pm$ 0.0 \\
Alarm      & 8.6 $\pm$ 2.24   & 11 $\pm$ 0.0 \\
Hailfinder & 47.6 $\pm$ 3.04  & 46 $\pm$ 0.0 \\
Win95pts   & 45.6 $\pm$ 5.04  & 44 $\pm$ 0.0 \\
Andes      & 158.2 $\pm$ 31.76 & 169 $\pm$ 0.0 \\
Pigs       & 20 $\pm$ 0.0     & 20 $\pm$ 0.0 \\
\bottomrule
\end{tabular}
\label{random_seeds} 
\end{table}

\subsection{DAG check table}
\label{sec:DAG_check}
To assess whether the learned graphs are already acyclic before post-processing, we report the DAG validity status and the number of edges removed during cycle projection. Tables~\ref{cycle_random} and~\ref{cycle_llm} summarize the statistics under random and LLM priors, respectively.

\begin{table}[t]
\centering
\small
\caption{Cycle projection statistics before DAG check(Random prior).}
\label{cycle_random}
\begin{tabular}{lccc}
\toprule
Dataset & DAG before check & Edges removed & Removed (\%) \\
\midrule
Sachs      & NO & 2 & 6\% \\
Child      & NO & 1 & 4\% \\
Insurance  & YES & 0 & 0 \\
Alarm      & YES & 0 & 0 \\
Hailfinder & YES & 0 & 0 \\
Win95pts   & YES & 0 & 0 \\
Andes      & NO & 6 & 2\% \\
Pigs       & YES & 0 & 0 \\
Link       & YES & 0 & 0 \\
\bottomrule
\end{tabular}
\end{table}

\begin{table}[t]
\centering
\small
\caption{Cycle projection statistics before DAG check(LLM prior).}
\label{cycle_llm}
\begin{tabular}{lccc}
\toprule
Dataset & DAG before check & Edges removed & Removed (\%) \\
\midrule
Sachs      & NO & 2 & 6\% \\
Child      & NO & 2 & 8\% \\
Insurance  & NO & 4 & 10\% \\
Alarm      & YES & 0 & 0 \\
Hailfinder & NO & 2 & 6\% \\
Win95pts   & NO & 5 & 6\% \\
Andes      & NO & 20 & 6\% \\
Pigs       & YES & 0 & 0 \\
Link       & YES & 0 & 0 \\
\bottomrule
\end{tabular}
\end{table}

\subsection{Runtime}
\label{sec:runtime}

Table ~\ref{FCI_runtime} represents the runtime of FCI/RFCI for all datasets.

\begin{table}[t]
\centering
\small
\caption{Runtime of FCI/RFCI and refinement.}
\label{FCI_runtime}
\begin{tabular}{lrrrr}
\toprule
Dataset & $|V|$ & sample size & FCI/RFCI (s) & Refinement (s) \\
\midrule
Sachs      & 11 & 855 & 0.98 & 2.8  \\
Child      & 20 & 5000 & 8.02 & 2.6  \\
Insurance  & 27 & 10000 & 9.69 & 38   \\
Alarm      & 37 & 10000 & 8.98 & 32   \\
hailfinder & 56 & 10000 & 4.20 & 59     \\
Win95pts   & 76 & 10000 & 8.64 & 71   \\
Andes      & 223& 10000 & 56.4 & 189  \\
Pigs       & 441& 50000 & 386.26 & 368  \\
Link       & 724& 50000 & 538.40 & 742  \\
\bottomrule
\end{tabular}
\end{table}

\end{document}